\documentclass[11pt,letter,onecolumn]{article}

\usepackage{solarized-light}
\usepackage{marginnote}
\usepackage{graphicx}
\usepackage{authblk,etoolbox}
\usepackage{titlesec}
\usepackage{calc}
\usepackage{tikz}
\usepackage{hyperref}
\usepackage{caption}
\usepackage{tcolorbox}
\usepackage{amssymb,amsmath}
\usepackage{ifxetex,ifluatex}
\usepackage{seqsplit}
\usepackage{color}
\pagecolor{white}
\usepackage{fixltx2e} % provides \textsubscript
\usepackage{natbib}
\bibliography{paper}
\usepackage{newfloat}
\DeclareFloatingEnvironment[
    fileext=los,
    listname={List of Snippets},
    name=Snippet,
    placement=tbhp,
    within=section,
]{snippet}

\usepackage[top=3.5cm, bottom=3cm, right=1.5cm, left=1.0cm,
            headheight=2.2cm, reversemp, includemp, marginparwidth=4.5cm]{geometry}

\titleformat{\section}
  {\normalfont\sffamily\Large\bfseries}
  {}{0pt}{}
\titleformat{\subsection}
  {\normalfont\sffamily\large\bfseries}
  {}{0pt}{}
\titleformat{\subsubsection}
  {\normalfont\sffamily\bfseries}
  {}{0pt}{}
\titleformat*{\paragraph}
  {\sffamily\normalsize}

\usepackage{fancyhdr}
\makeatletter
\let\ps@plain\ps@fancy
\fancyheadoffset[L]{4.5cm}
\fancyfootoffset[L]{4.5cm}

\definecolor{linky}{rgb}{0.0, 0.5, 1.0}

\newtcolorbox{repobox}
   {colback=red, colframe=red!75!black,
     boxrule=0.5pt, arc=2pt, left=6pt, right=6pt, top=3pt, bottom=3pt}

\newcommand{\ExternalLink}{%
   \tikz[x=1.2ex, y=1.2ex, baseline=-0.05ex]{%
       \begin{scope}[x=1ex, y=1ex]
           \clip (-0.1,-0.1)
               --++ (-0, 1.2)
               --++ (0.6, 0)
               --++ (0, -0.6)
               --++ (0.6, 0)
               --++ (0, -1);
           \path[draw,
               line width = 0.5,
               rounded corners=0.5]
               (0,0) rectangle (1,1);
       \end{scope}
       \path[draw, line width = 0.5] (0.5, 0.5)
           -- (1, 1);
       \path[draw, line width = 0.5] (0.6, 1)
           -- (1, 1) -- (1, 0.6);
       }
   }

\patchcmd{\@maketitle}{center}{flushleft}{}{}
\patchcmd{\@maketitle}{center}{flushleft}{}{}
\patchcmd{\@maketitle}{\LARGE}{\LARGE\sffamily}{}{}
\def\maketitle{{%
  
  \AB@maketitle}}
\makeatletter
\renewcommand\AB@affilsepx{ \protect\Affilfont}
\renewcommand\AB@affilnote[1]{{\bfseries #1}\hspace{3pt}}
\makeatother

\renewcommand\Affilfont{\sffamily\small\mdseries}
\ifnum 0\ifxetex 1\fi\ifluatex 1\fi=0 % if pdftex
  \usepackage[T1]{fontenc}
  \usepackage[utf8]{inputenc}

\else % if luatex or xelatex
  \ifxetex
    \usepackage{mathspec}
  \else
    \usepackage{fontspec}
  \fi
  \defaultfontfeatures{Ligatures=TeX,Scale=MatchLowercase}

\fi
\IfFileExists{upquote.sty}{\usepackage{upquote}}{}
\IfFileExists{microtype.sty}{%
\usepackage{microtype}
\UseMicrotypeSet[protrusion]{basicmath} % disable protrusion for tt fonts
}{}

\usepackage{hyperref}
\hypersetup{unicode=true,
            pdftitle={learn2learn: A Library for Meta-Learning Research},
            pdfborder={0 0 0},
            breaklinks=true}
\IfFileExists{parskip.sty}{%
\usepackage{parskip}
}{% else
\setlength{\parindent}{0pt}
\setlength{\parskip}{6pt plus 2pt minus 1pt}
}
\ifx\paragraph\undefined\else
\let\oldparagraph\paragraph
\renewcommand{\paragraph}[1]{\oldparagraph{#1}\mbox{}}
\fi
\ifx\subparagraph\undefined\else
\let\oldsubparagraph\subparagraph
\renewcommand{\subparagraph}[1]{\oldsubparagraph{#1}\mbox{}}
\fi

\title{\texttt{learn2learn}: A Library for Meta-Learning Research}

        \author[1]{Sébastien M. R. Arnold\footnote{Correspondance to
  \texttt{seb.arnold@usc.edu}}}
          \author[2]{Praateek Mahajan}
          \author[3]{Debajyoti Datta}
          \author[4]{Ian Bunner}
          \author[5]{Konstantinos Saitas Zarkias}
    
      \affil[1]{University of Southern California}
      \affil[2]{Iterable, Inc.}
      \affil[3]{University of Virginia}
      \affil[3]{University of Waterloo}
      \affil[5]{KTH Royal Institute of Technology and RISE - Research Institutes of
Sweden, SICS}
  \date{\vspace{-5ex}}

\begin{document}
\maketitle

\marginpar{
  %\hrule
  \sffamily\small

  {\bfseries Website:} \href{http://learn2learn.net}{\color{linky}{http://learn2learn.net}}

  \vspace{2mm}

  {\bfseries Software}
  \begin{itemize}
    \setlength\itemsep{0em}
    \item \href{http://github.com/learnables/learn2learn}{\color{linky}{Repository}} \ExternalLink
    \item \href{http://learn2learn.net/docs/learn2learn/}{\color{linky}{Documentation}} \ExternalLink
    \item \href{http://learn2learn.net/tutorials/getting\_started/}{\color{linky}{Tutorials}} \ExternalLink
  \end{itemize}

  \vspace{2mm}
  {\bfseries Licence}\\
  Authors of this paper retain copyright and release the work under a Creative Commons Attribution 4.0 International License (\href{http://creativecommons.org/licenses/by/4.0/}{\color{linky}{CC-BY}}).
}

\begin{abstract}
Meta-learning researchers face two fundamental issues in their empirical work: prototyping and reproducibility.
Researchers are prone to make mistakes when prototyping new algorithms and tasks because modern meta-learning methods rely on unconventional functionalities of machine learning frameworks.
In turn, reproducing existing results becomes a tedious endeavour -- a situation exacerbated by the lack of standardized implementations and benchmarks.
As a result, researchers spend inordinate amounts of time on implementing software rather than understanding and developing new ideas.

This manuscript introduces \texttt{learn2learn}, a library for meta-learning research focused on solving those prototyping and reproducibility issues.
\texttt{learn2learn} provides low-level routines common across a wide-range of meta-learning techniques 
(e.g. meta-descent, meta-reinforcement learning, few-shot learning),
and builds standardized interfaces to algorithms and benchmarks on top of them.
In releasing \texttt{learn2learn} under a free and open source license, we hope to foster a community around standardized software for meta-learning research.
\end{abstract}

\section{Introduction}\label{introduction}

Meta-learning is the subfield of machine learning that endows computer
programs with the ability of learning to learn. That is, the computer
not only learns a behavior, but also how to adapt its behavior. To
illustrate this difference, let us make an anology with the world of
athleticism. If a learning program is akin to an athlete, a
meta-learning program corresponds to the athlete-coach pair: it
simultaneously learns a skill and how to teach it best. Meta-learning is
appealing whenever the athlete is required to excel in multiple sports,
as the coach can leverage shared aspects of each sport to accelerate
mastery. Recent meta-learning methods emerged from this natural ability
to multitask -- e.g.~meta-descent in optimization, meta-reinforcement
learning in reinforcement learning, and few-shot learning when labelled
data is limited -- reaching state-of-the-art performance on vision,
language, and robotic domains. (Sutton 1992; Duan, Schulman, et al.
2016; Miller, Matsakis, and Viola 2000; Lee et al. 2019; Brown et al.
2020; Metz et al. 2019; Nagabandi et al. 2018)

Unfortunately, modern meta-learning research is often slowed down by
prototyping and reproducibility challenges. Prototyping algorithms is an
error-prone process since meta-learning algorithms rely on supported but
exotic functionalities of machine learning software. (e.g.~gradient of
optimization steps) Consequently, a slow prototyping phase reduces the
number of ideas one can try and retain. It also directly impacts
reproducibility: researchers are more likely to make mistake with
someone else's idea than with their own. Combined, those pesky issues
prevent meaningful comparisons across publications, ultimately reducing
the impact of any publication in the field.

\emph{Why do we suffer prototyping and reproducibility issues?}

Although a complete answer is outside the scope of this manuscript, we
blame the lack of specialized software as a major culprit. A software
library with specialized subroutines would reduce gaffes when
prototyping, improving reproducibility too. A widely-adopted library
also promotes standardized implementation of existing methods and
benchmarks, an issue in prior research; for example, the community lost
the original mini-ImageNet data splits from Vinyals et al. (2016),
leaving subsequent work to replicate them as best they could. In
summary, there is a need for a specialized software library which would
alleviate many of the issues currently plaguing meta-learning research.

We introduce \texttt{learn2learn}, a software library that directly
addresses prototyping and reproductibility issues in meta-learning
research. \texttt{learn2learn} provides researchers with a unified and
extensible interface to existing benchmarks, and a set of well-tested
subroutines frequently used to implement meta-learning algorithms. The
library also packages numerous examples replicating published methods,
which can easily be adapted for comparisons on new benchmarks.
\texttt{learn2learn} is implemented in Python to maintain compatiblity
with the greater machine learning ecosystem. It extends PyTorch (Paszke
et al. 2019) and leverages its fast linear algebra and automatic
differentiation capabilites, while resorting to Cython (Behnel et al.
2011) when speed is required for data-handling. \texttt{learn2learn} is
already used in our day-to-day research, and we hope its development
will continue to benefit the wider meta-learning community.

The remainder of this document overviews the prototyping and
reproducibility capabilities of \texttt{learn2learn}, including a review
of related work.

\section{Prototyping}\label{prototyping}

Prototyping is essential in letting researchers quickly try new ideas.
The faster the prototyping phase, the faster an idea can be retained (or
discarded) for further exploration. \texttt{learn2learn} provides tools
to accelerate two aspects of the prototyping phase: algorithms and
domains.

\paragraph{Algorithms}

Implementing meta-learning algorithms can be tricky. For example, many
methods rely on computing gradients of algorithms -- rather than
gradients of functions -- which, while possible with modern machine
learning frameworks (e.g.~PyTorch, TensorFlow, JAX), is strenuous and
prone to errors. (Finn, Abbeel, and Levine 2017; Jacobsen et al. 2019;
Xu, Hasselt, and Silver 2018) However, this doesn't need be the case:
many differentiable algorithms can be implemented with minor changes
when given the right abstractions.\footnote{See for example Agrawal et
  al. (2019).}

\begin{snippet}[h]
\begin{lstlisting}[language=Python]
learned_update = l2l.optim.ParameterUpdate(
    model.parameters(),
    l2l.optim.KroneckerTransform(l2l.nn.KroneckerLinear)
)
clone = l2l.clone_module(model)  # torch.clone() for nn.Modules
updates = learned_update(  # similar API as torch.autograd.grad
    loss(clone(X), y),
    clone.parameters(),
    create_graph=True,
)
# in-place, differentiable update of clone parameters
l2l.update_module(clone, updates)
# gradients w.r.t model's and learned_update's parameters
loss(clone(X), y).backward()
\end{lstlisting}
\caption{
Demonstration of differentiable optimization routines.
Lines 1-4 instantiate a parameterized update function, which computes the gradient of a loss w.r.t. some module's parameters (\texttt{clone}, line 8) and passes them through a \emph{gradient transform} -- a module mapping gradients to updates. (here a \texttt{KroneckerLinear}, line 3)
Line 5 creates a differentiable copy of the model, and line 11 updates that copy in-place such that the update is itself differentiable.
Finally, line 13 backpropagates through the loss of the updated differentiable copy, thus computing gradients w.r.t. the "pre-update" model parameters and the \texttt{KroneckerLinear} parameters.
\label{snip:diff-opt}
}
\end{snippet}

To ease the implementation of such methods, \texttt{learn2learn} exposes
low-level routines for differentiable optimization in
\texttt{learn2learn.optim}. These routines are tightly built around
PyTorch's automatic differentiation engine, so as to maintain
compatibility and extensibility. They can be used to express
optimization algorithms such that their computational graph remains
differentiable. Snippet \ref{snip:diff-opt} provides an example, which
implements the 1-step learning loop of the linear Kronecker-factored
optimizer from Arnold, Iqbal, and Sha (2019). (Note how both model and
optimizer parameters are meta-learned; implementing the same
functionality with vanilla PyTorch requires 10x the lines of code.)
We've used those general-purpose routines to implement algorithms from
the literatures of few-shot, meta-descent, and meta-reinforcement
learning -- including MAML (Finn, Abbeel, and Levine 2017),
Hypergradient descent (Baydin et al. 2017), or ProMP (Rothfuss et al.
2018) among others.

\paragraph{Domains}

Researchers have to design new domains -- such as datasets, tasks, or
environments -- to develop and test new abilities of their programs. We
refer to this other aspect of the research lifecycle as
\emph{prototyping new domains}. \texttt{learn2learn} can help prototype
new domains for few-shot and meta-reinforcement learning.

\begin{snippet}[h]
\begin{lstlisting}[language=Python]
dataset = l2l.data.MetaDataset(MyDataset())  # PyTorch dataset
transforms = [  # easy to define custom task transforms
    l2l.data.transforms.NWays(dataset, n=5),
    l2l.data.transforms.KShots(dataset, k=1),
    l2l.data.transforms.LoadData(dataset),
    lambda task: [(random_rotation(x), y) for x, y in task]
]
taskset = l2l.data.TaskDataset(dataset,
                               transforms,
                               num_tasks=20000)
random_task = taskset.sample()  # sample one task
for task in taskset:  # iterate over all tasks
    X, y = task
\end{lstlisting}
\caption{
    The interface to \texttt{TaskDataset} and \texttt{TaskTransform}.
    Line 1 wraps an arbitrary PyTorch dataset with the \texttt{MetaDataset} class.
    Lines 2-7 define \texttt{TaskTransforms}: here, 5-ways 1-shot classification with a custom random-rotation task augmentation applied to each (x, y) pair.
    Lines 8-10 instantiate the \texttt{TaskDataset}, from which tasks can be sampled (line 11) or enumerated (lines 12-13).
    \label{snip:task-dataset}
}
\end{snippet}

For few-shot meta-learning, \texttt{learn2learn} provides a general
\texttt{TaskDataset} class enabling sampling of smaller tasks in
\texttt{learn2learn.data}. Those tasks are constructed through a series
of \texttt{TaskTransform}s, which iteratively refine the description of
the data in the task. Writing a new task transform is as easy as writing
a Python function, but those functions can be made arbitrarily complex
thanks to Python's callable objects. See Snippet \ref{snip:task-dataset}
for an example. With the combination of \texttt{TaskDataset} and
\texttt{TaskTransform}s, researchers can quickly develop fast custom
data and task sampling schemes, while retaining compatibility with any
PyTorch dataset; this lets them quickly iterate over ideas with small
datasets and scale up to larger experiments with the same codebase.

\begin{snippet}[h]
\begin{lstlisting}[language=Python]
def make_env():
    env = l2l.gym.HalfCheetahForwardBackwardEnv()
    return cherry.envs.StateNormalizer(env)

# use 16 processes, compatible with gym API
env = l2l.gym.AsyncVectorEnv([make_env for i in range(16)])
tasks = env.sample_tasks(20)
env.set_task(tasks[0])  # all processes run task 0
\end{lstlisting}
\caption{
    Utilities for meta-reinforcement learning environments.
    Lines 1-3 instantiate the half-cheetah environment, with tasks defined as running forward or backward.
    This environment is then wrapped by cherry, an external reinforcement learning library.
    Line 6 forks 16 asynchronous workers, each with its own copy of the half-cheetah environment.
    Finally, lines 7-8 sample 20 tasks and assign the first one to all workers.
    \label{snip:meta-rl}
}
\end{snippet}

For meta-reinforcement learning, \texttt{learn2learn} provides a
high-level \texttt{MetaEnv} interface in \texttt{learn2learn.gym} which
can be used to bootstrap the design of OpenAI Gym environments.
(Brockman et al. 2016) Environments that adhere to this interface can
take advantage of specially designed utilities included in
\texttt{learn2learn}; for example, the \texttt{AsyncVectorEnv} wrapper
parallelizes the collection of episodes across multiple processes. (c.f.
Snippet \ref{snip:meta-rl}) Such environments also retain the Gym API,
making them compatible with all popular reinforcement learning
libraries. (Dhariwal et al. 2017; Duan, Chen, et al. 2016; Liang et al.
2017) Naturally, they also become compatible with the various
meta-reinforcement learning algorithms implemented within the library.

The core prototyping tools toured in the above paragraphs open the door
to more advanced developments such as online, incremental, or lifelong
meta-learning. While this manuscript can only present a bird's-eye view
of those tools, we invite the reader to the library's website and
documentation for such advanced applications.

\section{Reproducibility}\label{reproducibility}

The field of meta-learning is advancing rapidly, but progress is plagued
by reproducibility issues. Those issues are often subtle and hard to
spot. In meta-reinforcement learning for example, different papers have
used different reward functions with the same environment, resulting in
confusing comparisons: are the observed improvements due to algorithmic
advancements or to changes in the reward function? To combat those
insiduous issues, \texttt{learn2learn} includes a set of high-quality
implementations for various meta-learning algorithms as well as
standardized benchmarks for few-shot and meta-reinforcement learning.

\begin{snippet}[h]
\begin{lstlisting}[language=Python]
meta_sgd = l2l.algorithms.GBML(
    model,
    l2l.optim.ModuleTransform(l2l.nn.Scale),
)
meta_curvature = l2l.algorithms.GBML(
    model,
    l2l.optim.MetaCurvatureTransform,
)
meta_kfo = l2l.algorithms.GBML(
    model,
    l2l.optim.KroneckerTransform(l2l.nn.KroneckerLinear),
    adapt_transform=True,
)
\end{lstlisting}
\caption{
    Implementation of Meta-SGD (lines 1-4), Meta-Curvature (lines 5-8), and Meta-KFO (lines 9-13) with the \texttt{GBML} wrapper.
    Each variant differs in how the fast-adaptation gradients are transformed, which is reflected through the \texttt{transform} and \texttt{adapt\_transform} arguments.
    \label{snip:high-wrappers}
}
\end{snippet}

\paragraph{Implementations}

\texttt{learn2learn} provides high-level implementations for popular
algorithms. These implementations build on top of the low-level routines
from the previous section, and are thoroughly tested to replicate
published works. They typically wrap around PyTorch modules to extend
them with specific meta-learning functionalities. For example, the
\texttt{LearnableOptimizer} retains the familiar PyTorch
\texttt{Optimizer} interface and extends it to learn arbitrary
meta-optimization updates; similarly, the \texttt{GBML} augments PyTorch
\texttt{Module}s to support fast-adaptation routines for few-shot and
meta-reinforcement learning. \texttt{GBML} implementations of Meta-SGD,
Meta-Curvature, Meta-KFO are available in Snippet
\ref{snip:high-wrappers}. (Li et al. 2017; Park and Oliva 2019; Arnold,
Iqbal, and Sha 2019)

With those high-level implementations and the standardized benchmark
interface described below, we supply examples that exactly reproduce
published experiments. Those examples serve three purposes. First, they
validate the correctness of our implementation and a publication's
claims; second, they illustrate how to use the library; third, they fill
the need for unified standardized reproductions. As a by-product, those
examples can be used to bootstrap further experimentation and analysis
around a method of interest. For example, we could easily complement
ANIL's (Raghu et al. 2019) original results on the Omniglot (Lake,
Salakhutdinov, and Tenenbaum 2015) and mini-Imagenet (Vinyals et al.
2016) datasets with new results on CIFAR-FS (Bertinetto et al. 2018) and
FC100 (Oreshkin, Rodríguez López, and Lacoste 2018).

\begin{snippet}
\begin{lstlisting}[language=Python]
from learn2learn.vision import benchmarks
print(benchmarks.list_tasksets())
# ['omniglot', 'cifar-fs', 'fc100', 'mini-imagenet', ...]
tasksets = benchmarks.get_tasksets(  # standardized pipeline
    name='mini-imagenet',
    train_samples=10,
    train_ways=5)
task = tasksets.train.sample() # tasksets.train is a task dataset
\end{lstlisting}
\caption{
    High-level API to standardized computer vision benchmarks.
    Line 2 prints the list of available tasksets.
    On lines 4-7, we instantiate the 5-ways 5-shots mini-ImageNet benchmark, which returns the \texttt{tasksets} \texttt{namedtuple} with \texttt{train}, \texttt{validation}, and \texttt{test} keys.
    Line 8 samples a task from the train \texttt{TaskDataset}.
    \label{snip:high-benchmarks}
}
\end{snippet}

\paragraph{Benchmarks}

We use the low-level domains API to implement standard benchmarks in
few-shot and meta-reinforcement learning settings. For few-shot
learning, \texttt{learn2learn} provides classes to download and
preprocess datasets commonly used by the community in
\texttt{learn2learn.vision}. In addition, it also includes task
definitions with the proper task-processing stages (image normalization,
rotation, cropping) for commonly reported settings such as 5-ways
1-shot, 5-ways 5-shots, and 20-ways 5-shots. An example of those task
definitions is described in Snippet \ref{snip:high-benchmarks}.

The meta-reinforcement learning environments range from simple
2D-particle navigation to robotics control. In particular, we include
simple to use wrappers for the recently proposed MetaWorld, a set of 50
gripper manipulation tasks with varying levels of difficulty. Those
benchmark implementations should greatly simplify the replication and
comparison of new methods on well-studied settings. They, and other
meta-reinforcement learning utilities, are included in
\texttt{learn2learn.gym}.

Combined with the provided implementations, we hope \texttt{learn2learn}
enables researchers to accelerate the process of correctly comparing
their ideas against existing methods.

\section{Related Work}\label{related-work}

Disparate implementations of individual algorithms set aside, two recent
libraries tackle similar challenges as \texttt{learn2learn}.

The first one, \texttt{higher} (Grefenstette et al. 2019), aims to
facilitate the implementation of ``generalized inner-loop
meta-algorithms'' -- in other words, the implementation of
differentiable optimization algorithms. \texttt{higher} treats a model
definition (e.g.~a neural network) as a symbolic computational graph,
for which they use one set of parameters or another based on
user-specified context. This ``stateless'' parameterization is as
expressive as the \texttt{learn2learn.optim} submodule -- after all,
both are implemented on top of PyTorch -- but requires researchers to
carefully understand when they are working with symbolic or declarative
parts of their computation. Instead, \texttt{learn2learn} sticks with a
stateful and declarative style, already familiar to the PyTorch research
userbase. Moreover, \texttt{higher} completely forgoes reproducibility
issues as its focus is on the implementation of novel algorithms.

The second one, \texttt{Torchmeta} (Deleu et al. 2019), intends to
provide a unified interface to popular datasets, including classes to
easily download and process them. Specifically, it focuses on
standardized few-shot computer vision tasks, allowing researchers to
easily swap one dataset for another. However, supporting new datasets
with \texttt{Torchmeta} requires implementing a bridging class, even if
the dataset is already in standard PyTorch format. On the other hand,
\texttt{learn2learn}'s \texttt{TaskDataset} explicitly avoids such
bridging classes, and is designed to be compatible with any PyTorch
dataset (including text, speech, and others). \texttt{Torchmeta} also
povides a thin algorithmic wrapper to demonstrate the usage of their
library with gradient-based meta-learning algorithms; but, this wrapper
is not compatible with the majority of PyTorch's layers, nor custom
modules implemented by researchers. In comparison,
\texttt{learn2learn}'s differentiable optimization submodule uniformly
handles all PyTorch \texttt{Module}s (including custom ones), making it
a more applicable research tool.

Overall, \texttt{learn2learn} offers a more general solution to the
prototyping and reproduciblity issues encountered in day-to-day
research. For example, at the time of this writing, neither of the above
libraries supports meta-descent or meta-reinforcement learning.

\section{Conclusion}\label{conclusion}

This manuscript introduces \texttt{learn2learn}, a library that tackles
prototyping and reproducibility issues in modern meta-learning.
\texttt{learn2learn}'s low-level routines facilitate rapid prototyping
of new algorithms and domains. The library builds on those low-level
routines to provide high-level implementations and standardized
benchmarks API, which enable researchers to easily and faithfully
compare different methods under different settings. Notably, both low-
and high-level utilities are engineered to be general: they are
compatible with a wide range of meta-learning techniques in few-shot
learning, meta-reinforcement learning, or meta-optimization. Finally,
\texttt{learn2learn} is released under the free and open source MIT
licence, and the focus of continued development.

\subsection{Acknowledgements}\label{acknowledgements}

We thank Fei Sha for providing excellent research environment and
guidance, and for the computational resources required to develop
\texttt{learn2learn}. We also thank the various users who -- through
their questions, comments, or contributions -- helped improve
\texttt{learn2learn}.

This work is partially supported by NSF IIS-1065243,1451412, 1513966/
1632803/1833137, 1208500, CCF-1139148,DARPA Award\#: FA8750-18-2-0117,
DARPA-D3M - AwardUCB-00009528, Google Research Awards, an Alfred P.
SloanResearch Fellowship, gifts from Facebook and Netflix, and
ARO\#W911NF-12-1-0241 and W911NF-15-1-0484.

\pagebreak

\section*{References}\label{references}
\addcontentsline{toc}{section}{References}

\hypertarget{refs}{}
\hypertarget{ref-cvxpylayers2019}{}
Agrawal, A., B. Amos, S. Barratt, S. Boyd, S. Diamond, and Z. Kolter.
2019. ``Differentiable Convex Optimization Layers.'' In \emph{Advances
in Neural Information Processing Systems}.

\hypertarget{ref-Arnold2019-co}{}
Arnold, Sébastien M R, Shariq Iqbal, and Fei Sha. 2019. ``When MAML Can
Adapt Fast and How to Assist When It Cannot,'' October.

\hypertarget{ref-Baydin2017-ws}{}
Baydin, Atilim Gunes, Robert Cornish, David Martinez Rubio, Mark
Schmidt, and Frank Wood. 2017. ``Online Learning Rate Adaptation with
Hypergradient Descent,'' March. 

\hypertarget{ref-behnel2011cython}{}
Behnel, Stefan, Robert Bradshaw, Craig Citro, Lisandro Dalcin, Dag
Sverre Seljebotn, and Kurt Smith. 2011. ``Cython: The Best of Both
Worlds.'' \emph{Computing in Science \& Engineering} 13 (2). IEEE:
31--39.

\hypertarget{ref-Bertinetto2018-gn}{}
Bertinetto, Luca, Joao F Henriques, Philip Torr, and Andrea Vedaldi.
2018. ``Meta-Learning with Differentiable Closed-Form Solvers.''

\hypertarget{ref-Brockman2016-bh}{}
Brockman, Greg, Vicki Cheung, Ludwig Pettersson, Jonas Schneider, John
Schulman, Jie Tang, and Wojciech Zaremba. 2016. ``OpenAI Gym,'' June.

\hypertarget{ref-Brown2020-zg}{}
Brown, Tom B, Benjamin Mann, Nick Ryder, Melanie Subbiah, Jared Kaplan,
Prafulla Dhariwal, Arvind Neelakantan, et al. 2020. ``Language Models
Are Few-Shot Learners,'' May. 

\hypertarget{ref-Deleu2019-yd}{}
Deleu, Tristan, Tobias Würfl, Mandana Samiei, Joseph Paul Cohen, and
Yoshua Bengio. 2019. ``Torchmeta: A Meta-Learning Library for PyTorch,''
September. 

\hypertarget{ref-baselines}{}
Dhariwal, Prafulla, Christopher Hesse, Oleg Klimov, Alex Nichol,
Matthias Plappert, Alec Radford, John Schulman, Szymon Sidor, Yuhuai Wu,
and Peter Zhokhov. 2017. ``OpenAI Baselines.'' \emph{GitHub Repository}.

\hypertarget{ref-Duan2016-iq}{}
Duan, Yan, Xi Chen, Rein Houthooft, John Schulman, and Pieter Abbeel.
2016. ``Benchmarking Deep Reinforcement Learning for Continuous
Control,'' April. 

\hypertarget{ref-Duan2016-pw}{}
Duan, Yan, John Schulman, Xi Chen, Peter L Bartlett, Ilya Sutskever, and
Pieter Abbeel. 2016. ``RL\(^2\): Fast Reinforcement Learning via Slow
Reinforcement Learning,'' November.

\hypertarget{ref-Finn2017-gw}{}
Finn, Chelsea, Pieter Abbeel, and Sergey Levine. 2017. ``Model-Agnostic
Meta-Learning for Fast Adaptation of Deep Networks.'' \emph{arXiv
Preprint arXiv:1703. 03400}.

\hypertarget{ref-Grefenstette2019-zs}{}
Grefenstette, Edward, Brandon Amos, Denis Yarats, Phu Mon Htut, Artem
Molchanov, Franziska Meier, Douwe Kiela, Kyunghyun Cho, and Soumith
Chintala. 2019. ``Generalized Inner Loop Meta-Learning,'' October.

\hypertarget{ref-Jacobsen2019-gd}{}
Jacobsen, Andrew, Matthew Schlegel, Cameron Linke, Thomas Degris, Adam
White, and Martha White. 2019. ``Meta-Descent for Online, Continual
Prediction,'' July.

\hypertarget{ref-Lake2015-uv}{}
Lake, Brenden M, Ruslan Salakhutdinov, and Joshua B Tenenbaum. 2015.
``Human-Level Concept Learning Through Probabilistic Program
Induction.'' \emph{Science} 350 (6266): 1332--8.

\hypertarget{ref-Lee2019-cl}{}
Lee, Kwonjoon, Subhransu Maji, Avinash Ravichandran, and Stefano Soatto.
2019. ``Meta-Learning with Differentiable Convex Optimization,'' April.

\hypertarget{ref-Li2017-li}{}
Li, Zhenguo, Fengwei Zhou, Fei Chen, and Hang Li. 2017. ``Meta-SGD:
Learning to Learn Quickly for Few-Shot Learning,'' July.

\hypertarget{ref-Liang2017-gx}{}
Liang, Eric, Richard Liaw, Philipp Moritz, Robert Nishihara, Roy Fox,
Ken Goldberg, Joseph E Gonzalez, Michael I Jordan, and Ion Stoica. 2017.
``RLlib: Abstractions for Distributed Reinforcement Learning,''
December. 

\hypertarget{ref-Metz2019-rf}{}
Metz, Luke, Niru Maheswaranathan, Jonathon Shlens, Jascha
Sohl-Dickstein, and Ekin D Cubuk. 2019. ``Using Learned Optimizers to
Make Models Robust to Input Noise,'' June.

\hypertarget{ref-Miller2000-hy}{}
Miller, E G, N E Matsakis, and P A Viola. 2000. ``Learning from One
Example Through Shared Densities on Transforms.'' In \emph{Proceedings
IEEE Conference on Computer Vision and Pattern Recognition. CVPR 2000
(Cat. No.PR00662)}, 1:464--71 vol.1.

\hypertarget{ref-Nagabandi2018-cz}{}
Nagabandi, Anusha, Ignasi Clavera, Simin Liu, Ronald S Fearing, Pieter
Abbeel, Sergey Levine, and Chelsea Finn. 2018. ``Learning to Adapt in
Dynamic, Real-World Environments Through Meta-Reinforcement Learning,''
March. 

\hypertarget{ref-Oreshkin2018-yi}{}
Oreshkin, Boris, Pau Rodríguez López, and Alexandre Lacoste. 2018.
``TADAM: Task Dependent Adaptive Metric for Improved Few-Shot
Learning.'' In \emph{Advances in Neural Information Processing Systems
31}, edited by S Bengio, H Wallach, H Larochelle, K Grauman, N
Cesa-Bianchi, and R Garnett, 721--31. Curran Associates, Inc.

\hypertarget{ref-Park2019-or}{}
Park, Eunbyung, and Junier B Oliva. 2019. ``Meta-Curvature,'' February.

\hypertarget{ref-Paszke2019-fx}{}
Paszke, Adam, Sam Gross, Francisco Massa, Adam Lerer, James Bradbury,
Gregory Chanan, Trevor Killeen, et al. 2019. ``PyTorch: An Imperative
Style, High-Performance Deep Learning Library.'' In \emph{Advances in
Neural Information Processing Systems 32}, edited by H Wallach, H
Larochelle, A Beygelzimer, F d'Alché-Buc, E Fox, and R Garnett,
8026--37. Curran Associates, Inc.

\hypertarget{ref-Raghu2019-ff}{}
Raghu, Aniruddh, Maithra Raghu, Samy Bengio, and Oriol Vinyals. 2019.
``Rapid Learning or Feature Reuse? Towards Understanding the
Effectiveness of MAML,'' September.

\hypertarget{ref-Rothfuss2018-dx}{}
Rothfuss, Jonas, Dennis Lee, Ignasi Clavera, Tamim Asfour, and Pieter
Abbeel. 2018. ``ProMP: Proximal Meta-Policy Search.''

\hypertarget{ref-Sutton1992-ar}{}
Sutton, Richard S. 1992. ``Adapting Bias by Gradient Descent: An
Incremental Version of Delta-Bar-Delta,'' June.

\hypertarget{ref-Vinyals2016-cm}{}
Vinyals, Oriol, Charles Blundell, Timothy Lillicrap, Koray Kavukcuoglu,
and Daan Wierstra. 2016. ``Matching Networks for One Shot Learning.'' In
\emph{Advances in Neural Information Processing Systems 29}, edited by D
D Lee, M Sugiyama, U V Luxburg, I Guyon, and R Garnett, 3630--8. Curran
Associates, Inc.

\hypertarget{ref-Xu2018-zh}{}
Xu, Zhongwen, Hado van Hasselt, and David Silver. 2018. ``Meta-Gradient
Reinforcement Learning,'' May. 

\bibliography{paper}

\end{document}